\pdfoutput=1

\documentclass[11pt]{article}

\usepackage{acl}

\usepackage{times}
\usepackage{latexsym}

\usepackage[T1]{fontenc}

\usepackage[utf8]{inputenc}

\usepackage{microtype}

\usepackage{inconsolata}

\usepackage{xspace}
\usepackage{amsmath}
\usepackage[compact]{titlesec}

\usepackage{graphicx}

\usepackage{tabularx}
\usepackage{adjustbox}

\newcommand{\tbf}[1]{\textbf{#1}}
\newcommand{\doublespace}[1]{\xspace\xspace}
\newcommand{\singlespace}[1]{\xspace}

\newcommand{\hartfrz}[0]{HaRT{\tiny{frozen}}\xspace}
\newcommand{\hartconcat}[0]{HaRT{\tiny{concat}}\xspace}
\newcommand{\hartonedoc}[0]{HaRT{\tiny{ODPB}}\xspace}
\newcommand{\ltinsep}[0]{LT{\tiny{insep}}\xspace}

\title{Evaluation of LLMs-based Hidden States as Author Representations for Psychological Human-Centered NLP Tasks}

\author{Nikita Soni, Pranav Chitale, Khushboo Singh, \\ \textbf{Niranjan Balasubramanian, H. Andrew Schwartz} \\
Stony Brook University\\
\texttt{\{nisoni, pchitale, khusingh, niranjan, has\}@cs.stonybrook.edu}}

\begin{document}
\maketitle
\begin{abstract}



Like most of NLP, models for human-centered NLP tasks---tasks attempting to assess author-level information---predominantly use representations derived from hidden states of Transformer-based LLMs. 
However, what component of the LM is used for the representation varies widely. Moreover, there is a need for Human Language Models (HuLMs) that implicitly model the author and provide a user\footnote{`Author', 'user' and `person' used interchangeably.}-level hidden state. 
Here, we systematically evaluate different ways of representing documents and users using different LM and HuLM architectures to predict task outcomes as both dynamically changing states and averaged trait-like user-level attributes of \textit{valence}, \textit{arousal}, \textit{empathy}, and \textit{distress}. 
We find that representing documents as an average of the token hidden states performs the best generally. 
Further, while a user-level hidden state itself is rarely the best representation, we find its inclusion in the model strengthens token or document embeddings used to derive document- and user-level representations resulting in best performances.



%

\end{abstract}

\section{Introduction}
\label{sec:intro}

Human-centered NLP focuses on the humans generating language~\citep{flek_returning_2020, hovy_importance_2021, soni2024proceedings} 
Many human-centered NLP tasks focus on assessing human-attributes of a user based on their language~\cite{lynn_human_2017}, utilizing a representation of the author, often within a specific time period or set of documents~\cite{matero-etal-2022-understanding,giorgi2024findings}. 


Like most sub-fields in NLP, human-centered NLP heavily relies on traditional LLMs (pre-trained on the language modeling task of either next word prediction or the missing word prediction) that do not 
directly contain a representation for the person. For example, no particular layer, token, or output is accepted as the best way to represent a person. 
More recently, \textit{human language models} (HuLMs)---those that directly model the language in the context of its ``generator'', i.e., the author ~\cite{soni2024large}---have been proposed but still lack evaluations for capturing human-level factors.


\begin{figure}[tb!]
    \centering

  \includegraphics[width=\linewidth]{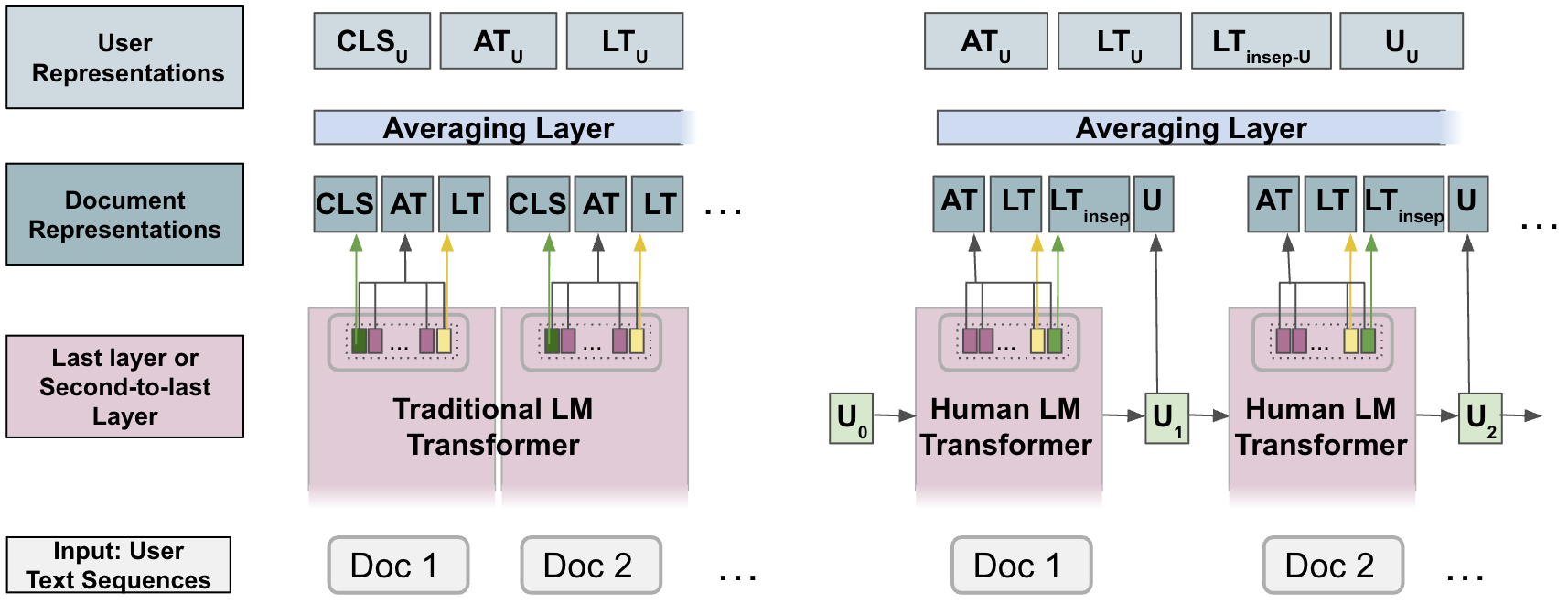}  
  \caption{Evaluating representations derived from different hidden states in both traditional Transformer-based LMs and human language models (HuLMs). Document representations as hidden states from: CLS, special token in auto-encoders; AT, averaged tokens in a document; LT, last token in a document for autoregressive models; \ltinsep, special last token for HuLMs; and U, user state for HuLMs. User representation as averaged corresponding document representation types for all documents written by a user.}
  \label{reps_fig}    
\end{figure}

Here, we systematically compare and evaluate the commonly used ways to represent \textbf{words, documents, and users} in terms of hidden states (i.e., vector embeddings) from traditional LMs and HuLMs (Figure \ref{reps_fig}) for their ability to capture user-level information. We use the three levels of representations (word, document, user) to predict three levels of task outcomes: \textbf{document-level, wave-level, and user-level}. These outcomes are selected based on the prediction properties of: a) changing user states – as an effect of different situations/times and particular states when an author writes the document, b) changing user states over a range of (closely spaced) periods of time --- as an effect of differences in user’s states grouped together in one period of time (referring to it as a wave) and specifically separated from the next period of time (i.e., next wave), and c) stable trait-like user-level attributes. We select four well-established outcomes representative of psychological human-centered NLP tasks spanning fundamental human attributes: the affective circumplex~\cite{Russell1980, mohammad2016sentiment} (valence and arousal), empathy, and distress~\cite{Batson1987DistressAE, litvak2016social})

Transformer-based LM~\cite{vaswani2017attention} architectures provide us with many choices for document representation: first token (i.e., CLS)~\cite{devlin-etal-2019-bert, liu2019roberta} in case of auto-encoders, the last token in case of autoregressive models~\cite{radford2019language}, or the average of all the tokens in a document. Since traditional LMs do not have a notion of a user, prior works have used a hierarchical user representation by averaging all document embeddings written by a particular user \cite{nagao-katsurai-2024-researcher} (Figure \ref{reps_fig}). While that is a posthoc way to represent a user, HuLMs provide us with user representations -- a relatively more natural way for the same, by processing language in its multi-level structure i.e., words are a part of a document, and documents are generated by people~\cite{soni-etal-2022-human}. We select representative models for these different types of LMs currently available: autoencoders (BERT-like), autoregressive (GPT2-like), and HuLMs (HaRT-like \cite{soni-etal-2022-human}). HuLMs are trained on user-aware LM pre-training tasks that predict the next word conditioned on the contextual words \textit{as well as a user state}, thereby processing text in the context of their historical language by tracking \textit{user states} through recurrently updated vectors. HuLMs output word embeddings like traditional LMs (however, these are user-state-informed), and also output user states presenting additional opportunities for producing rich user representations from such user states. We hypothesized the user representations derived from user states---as the natural output of HuLMs---will result in better assessing the human-attributes considered at the three different levels of predictions. 

\paragraph{Contributions.}
(1) We depict some of the different ways to represent words, documents, and users using traditional LMs and HuLMs hidden states and user states to predict a user's frequently changing states (i.e., document-level), their changing states over a specific period of time (i.e., wave), and their stable trait-like attributes (i.e., user-level). 
(2) We systematically evaluate the multiple representation types over broadly 2 levels of analysis of user attributes: document- and wave-level dynamically changing over time (states) and somewhat stable user-level (trait-like) on an average. Traits are defined as averages of multiple state measurements (i.e. a trait defines the mean and sometimes the variance of one's distribution of states \cite{fleeson2001toward}).
(3) We draw 2 major findings: (a) averaging the token embeddings as the basis for document representations (states) and hierarchical user representations (traits) perform the best on an average, and (b) contrary to our hypothesis, user states themselves are not the best representations, but consistently strengthen user representations derived from token and document embeddings where a user-state was included.

\section{Related Work}

Human-centered NLP tasks are often grounded in the person \citep{ganesan2024text} and require user representations for downstream applications, such as assessing mental health~\cite{ganesan2022wwbp, varadarajan-etal-2024-archetypes},
estimating demographic attributes~\cite{benton-etal-2016-learning}, assessing personality~\cite{schwartz_personality_2013, soni2024comparing}, stance detection~\cite{matero-etal-2021-melt-message}, 
and personalized recommendations~\cite{hierec, wu2023personalized}. Commonly, a hierarchical approach has been used to average document representations across all text sequences from a particular user since the pre-LLMs era~\cite{hierarchicalselfattentive, lynn_hierarchical_2020}. As Transformer-based LLMs become ubiquitous in human-centered NLP tasks, the hierarchical user representation approach continues to prevail~\cite{lynn_hierarchical_2020, nagao-katsurai-2024-researcher}. While this allows us to derive a user representation, it does not necessarily capture the author's context, as traditional LLMs process documents written by the same individual independently or, conversely, treating all language as if written by an average, universal person~\cite{soni2024large}.

A large body of past work explored generating user embeddings~\citep{ning2024userllmefficientllmcontextualization} and making LLMs personalizable~\citep{zhang2024personalizationlargelanguagemodels}.
Recently introduced Transformer-based Human Language Models (HuLMs)~\cite{soni-etal-2022-human} offer a notion of the user by processing language in its multi-level structure, where words are part of a document, and documents are generated by people. This provides a relatively natural way of representing a user. Traditional LMs and HuLMs present a multitude of choices regarding which layer, token, or output to select for the best representation of a person and their document- and token-level representations. To this end, we perform a systematic study to identify the best representations.



\section{Datasets and Tasks}
\label{sec:data}

We select 2 datasets that allow for the multiple levels of analysis broadly categorized into (a) dynamic states (attributes that change with time and utterances), and (b) user-level (trait-like attributes on an average for a person). These datasets provide tasks of predicting fundamental human-attributes like affective circumplex grid \cite{Russell1980} (valence and arousal), and empathy and distress \cite{Batson1987DistressAE}.

\paragraph{Valence and Arousal.} We use a subset of the DS4UD (Data Science for Unhealthy Drinking) dataset~\cite{nilsson2024language}, consisting of affective circumplex grid self-reported questionnaires and essays written by US service industry workers on how they were feeling 3 times daily (minimum once) for 14-day periods (we call it a ``wave''), across 6 data collection phases from 2021 to early 2024. We use self-reported outcomes derived from the affective circumplex grid: \textit{Valence} (0-4, highly negative to highly positive affect), and \textit{Arousal} (0-2, low to high energy). Our selection criteria include users who participated in at least 2 waves and wrote at least 2 essays ($\geq$ 10 words each), resulting in 120 users overall, 406 user instances across 6 waves (a user can have 2 to 6 wave instances, depending on the number of waves they participated in), and 10090 essays across all users. We compute outcomes labels across waves by averaging the labels corresponding to each essay for a user in a particular wave. Further, we average the labels corresponding to each wave for a user to get the user-level outcomes labels, following classical test theory ~\cite{rust2014modern}.






\paragraph{Empathy and Distress.} We use a subset of the WASSA 2024 Shared Task on Empathy and Personality Detection in Interactions, specifically a part of Track 3 \cite{giorgi2024findings}. The task is to predict the empathic concern and personal distress scores \cite{Batson1987DistressAE} for each essay written by crowd workers in response to news articles, with real-valued labels ranging from 1 to 7. We include train and dev data from WASSA 2024 and WASSA 2023 \cite{barriere-etal-2023-findings}, the years with user-level data, resulting in 180 users and 1837 essays. User-level trait-like outcome labels are obtained by averaging empathy and distress scores across each user's essays.





\section{Methods and Experiments}
\label{sec:methods}




\paragraph*{Models and Types of Representations.}
In this study, we will evaluate different types of representations from 4 models, of which 2 are autoregressive: 1) HaRT \citep{soni-etal-2022-human}, 2) GPT-2 small, and 2 are auto-encoders: 3) BERT, 4) RoBERTa.

HaRT enables us to use 4 types of \textbf{document representations}: a) Average of token embeddings (AT), b) Last token (LT), c) Last token as \textit{insep} (\ltinsep), d) User states (U) (Figure \ref{reps_fig}). Autoregressive GPT-2 can provide the first 2 of these 4 types, while auto-encoders (BERT, RoBERTa) normally use the CLS token embeddings (CLS) or the AT to represent a document. We employ the hierarchical averaging structure over the different types of document embeddings across all documents from a user, or in a specific time-period (i.e., wave) to retrieve different types of \textbf{user and wave representations} (refer Section \ref{sec:data}), respectively.

\begin{table*}[h!]
    \centering
    \renewcommand{\arraystretch}{0.907} 
    \begin{small}
    \begin{adjustbox}{width=\textwidth}
    \begin{tabular}{l|ccccc|ccc|ccccc}
        
        \hline
        
        \tbf{} & \multicolumn{8}{c|}{\tbf{Dynamic States}} & \multicolumn{5}{c}{\tbf{Trait-like}} \\
        
        \tbf{Method} & \multicolumn{5}{c|}{\tbf{Document-Level}} & \multicolumn{3}{c|}{\tbf{Wave-Level}} & \multicolumn{5}{c}{\tbf{User-Level}} \\
        
         \tbf{} & \tbf{Val} & \tbf{Aro} & \tbf{Emp} & \tbf{Dis} & \tbf{Avg} & \tbf{Val} & \tbf{Aro} & \tbf{Avg} & \tbf{Val} & \tbf{Aro} & \tbf{Emp} & \tbf{Dis} & \tbf{Avg} \\
         
         \hline
         

         GPT-2 &&&&&&&&&&&&&\\
         
         \hspace{15pt} LT &0.53&0.25&0.71&0.58&0.52&0.64&0.28&0.46&
         0.63&-0.14&0.37&0.42&0.32 \\
         
         \hspace{15pt} AT &0.60&0.34&0.69&0.54&\text{\doublespace{}\singlespace{}{}}0.54*&0.77& \bf 0.33&\text{\doublespace{}\singlespace{}}\bf{0.55*}&
         0.75&0.18&0.52&\bf0.68&\text{\doublespace{}} 0.53* \\

         BERT &&&&&&&&&&&&&\\
         
         \hspace{15pt} CLS &0.58&0.30&0.70&0.66&0.56&0.74&0.28&0.51&
         0.71&0.09&0.45&0.42&0.42 \\
         
         \hspace{15pt} AT &0.61& 0.35&0.72&0.59&\text{\doublespace{}\singlespace{}}0.57*&0.75&0.26&0.50&
         0.73&0.14&0.45&0.39&0.43 \\

         RoBERTa &&&&&&&&&&&&&\\
         
         \hspace{15pt} CLS &0.62&0.35& \bf 0.77&0.63&\text{\doublespace{}\singlespace{}}0.59*&0.75&0.29&0.52&
         0.72&0.10&0.51&0.59&0.48 \\
         
         \hspace{15pt} AT & 0.63& \bf 0.36&0.72&0.63&0.58&0.75& \bf 0.33 &\text{\doublespace{}\doublespace{}\singlespace{}}0.54**& 0.75&0.16& 0.59&0.53 & 0.51 \\

         
         

         \hartfrz &&&&&&&&&&&&&\\
         
        \hspace{15pt} LT &0.52&0.22&0.70&0.57&0.50&0.69&0.21&0.45&
        0.73&0.18&0.42&0.60&0.48 \\
                 
        \hspace{15pt} \ltinsep &0.55&0.23& 0.76&0.67&0.55&0.69&0.22&0.46&
        0.74&0.23&0.45& 0.62&0.51 \\
        
        \hspace{15pt} AT & 0.64& 0.33&  0.75& 0.68 &\text{\doublespace{}\singlespace{}}0.60*& 0.79&0.27 &\text{\doublespace{}\singlespace{}}0.53*& 0.73& \bf 0.35& \bf 0.67&0.66& \text{\doublespace{}\doublespace{}} \bf{0.60*}$\dagger$ \\
        
        \hspace{15pt} U &-&-&-&-&-&-&-&-&0.59&-0.01&0.46&0.49&0.38 \\

        \hartconcat &&&&&&&&&&&&& \\
        
        \hspace{15pt} LT &0.53&0.22&0.72&0.62&0.52& 0.70&0.16&0.43&
        0.74&0.18&0.42& \bf 0.68&0.50 \\
        
        \hspace{15pt} \ltinsep &0.57&0.26& 0.76&0.56&0.54&0.75& 0.30&0.53& \bf 0.77&0.15&0.61&0.55&0.52 \\
        
        \hspace{15pt} AT &\bf 0.65& 0.34& 0.75& \bf 0.70&\text{\doublespace{}\doublespace{}} \bf 0.61*$\dagger$& \bf 0.80& 0.27&0.54& 
        \bf 0.77& 0.34&0.62&0.65& \text{\doublespace{}\singlespace{}}0.59* \\
        
        \hspace{15pt} U &-&-&-&-&-&-&-&-&0.65&0.06&0.59&0.49&0.45 \\
        
        \hartonedoc &&&&&&&&&&&&& \\
        
        \hspace{15pt} LT &0.57&0.27&0.68&0.56&0.52&0.72&0.17&0.44&
        0.72&0.24&0.41&0.55&0.48 \\
        
        \hspace{15pt} \ltinsep & 0.64& 0.34&0.74&0.66&0.59& 0.79&0.27&0.53&
        0.72&0.09&0.40&0.50&0.43 \\
        
        \hspace{15pt} AT & 0.63& \bf 0.36&0.73&0.63&\text{\doublespace{}\singlespace{}}0.59*& \bf 0.80& 0.28&0.54&
        0.75&0.12&0.58&0.51&\text{\doublespace{}\doublespace{}\singlespace{}}0.49** \\
        
        \hspace{15pt} U &0.52&0.15&0.46&0.40&0.38&0.65&0.01&0.33&
        0.66&0.04&0.50&0.47&0.42 \\
               \hline
    \end{tabular}
    \end{adjustbox}
    \end{small}
    \caption{
    Linear model predictions for 4 well-established psychological outcomes: valence, arousal, empathy, and distress, correlated (Pearson $r$) with the labels associated with the datasets in Section \ref{sec:data}. Averaged token embeddings (AT) generally outperform LT, CLS, \ltinsep, and U across respective models. Token and document embeddings informed by the user state perform best for averaged document- and user-level outcomes. No statistically significant difference found between \hartconcat and GPT-2 for averaged wave-level outcomes. Bold indicates best in column and * indicates paired t-test statistical significance with p < 0.001 and ** with p < 0.05 for each model's AT results against its respective second best result. $\dagger$ represents statistical significance (p < 0.05) across models for each of document-, wave-, and user-level analysis between best overall best (HaRT variants) and best traditional LM.}
    
    \label{tab:main_results}
\end{table*}

\paragraph{HaRT Architecture and Variants.}
We sought to use an architecture that can handle multiple representations criteria: (1) a multi-level structure spanning words, documents, and user-level data, (2) last-token and special-token representations, and (3) applicability to traditional LM tasks. 
For our primary results we build on Human-aware Recurrent Transformer (HaRT) model that was pre-trained on the human language modeling task~\citep{soni-etal-2022-human} (see Section \ref{sec:intro}). Originally, HaRT processes temporally ordered text sequences from a user, separated by a special \textit{insep} token, chunked into blocks (up to 8 for training), each capped at 1024 tokens (referred to \textbf{\hartfrz}). HaRT maintains user state vectors, updated recurrently after processing each block. However, concatenated blocks of 1024 tokens didn’t provide an easy way to extract user states at the end of each document for document- and wave-level prediction. To address this, we further pre-train HaRT on DS4UD essays data (refer Section \ref{sec:data}) with a limit of one document per block (referred to \textbf{\hartonedoc}) and adjusting the maximum number of training blocks (to 55) to retain comparable user-level data, enabling user state extraction after each document. Additionally, in another variant (\textbf{\hartconcat}), we continue to pre-train \hartfrz using the DS4UD essays data with the original block-settings to help tease out any spurious effects introduced bt altered block-settings in \hartonedoc. Following \citeauthor{soni-etal-2022-human}, we derive user and wave representations by averaging user states across blocks in each HaRT variant (see Section \ref{sec:data}).




\paragraph{Experiments.}
We use averaged token embeddings, special token embeddings, and user states embeddings (where available) from HaRT variants, GPT-2, BERT, and RoBERTa\footnote{Code available at \href{https://github.com/soni-n/LLMs_Author_Representations}{https://github.com/soni-n/LLMs\_Author\_Representations}} to predict valence, arousal, empathy, and distress as both dynamic states and trait-like stable outcomes based on a person's text sequences (Table \ref{tab:main_results}). 
We use DLATK~\citep{DLATKemnlp2017} to apply ridge regression (aplha values: 1.e+00, 1.e+01, 1.e+02, 1.e+03, 1.e+04, 1.e+05, 1.e+06) over the different embeddings representations, learning only the ridge regression weights without any task-specific fine-tuning model parameters. 

To evaluate vector representations, we measure Pearson $r$ correlations between the linear model predictions and dataset labels (Section \ref{sec:data}). We run 10- straitified group-based cross-validation folds, ensuring no user overlap between the train and test splits in any fold, and report the averaged Pearson $r$. The experiments took \textasciitilde5 hours on A6000 GPU using approx \textasciitilde24GB memory.


\section{Results and Discussion}

Overall, averaged across outcomes, we find HuLMs (HaRT variants) to perform the best for predicting both the changing states and trait-like outcomes at the document- and user-level (refer Table \ref{tab:main_results}). We do not find any statistically significant difference in the performance of GPT-2 and \hartconcat for wave-level predictions.

\paragraph{Averaged Token Embeddings (AT).} Using the averaged token embeddings helps more often than the last token or CLS embeddings for all types of models considered: autoregresive, auto-encoders, and HuLMs. Intuitively, AT embeddings perform better because they may be better able to capture the nuanced, context-dependent details needed to represent both stable traits and changing states---key aspects of human-level attributes---from each token of a document. Whereas, when we use a single special token’s embeddings as a document representation, we may lose important nuanced information distributed across the tokens in the document in the process of compressing information into one token’s embeddings. However, this is only theoretical and we do not have evidence to prove the reasoning in the current study and encourage future studies to explore further.



\paragraph{Inclusion of User States.} The user states by themselves are rarely the best embeddings but consistently the LT and AT-based embeddings were strong when a user-state was included. We see under the HuLMs (HaRT variants) in Table \ref{tab:main_results}, representations from U to perform below par, however user-level representations derived from AT (and in some cases last tokens LT and \ltinsep), that are processed in the context of the user states, boost the performance. While using AT is the current best choice, this also opens up avenues for future works to improve the natural representations of user states capturing the author's context and thus further improve the word's meaning embedded in tokens.

\paragraph{Last Token as a Special Token (\ltinsep).} Interestingly, we find representations derived from \ltinsep embeddings from HuLMs (HaRT variants) to perform better than those from LT embeddings (Table \ref{tab:main_results}) that suggests \textit{insep} is able to capture the average of the tokens in a documents from a user. Theoretically, \ltinsep, the special token for HuLMs, can be seen as analogous to the CLS token for auto-encoders.

\paragraph{Outcomes Predictions Analysis.} Arousal prediction remains a challenging task across all 3 different levels of predictions (document, wave, and user). However, RoBERTa predicts it best at the document- and wave-level, while \hartfrz is the best in predicting it as an average trait for the author (i.e., user-level).

Valence is better predicted at the user- and wave-level across models, suggesting it is harder to capture the frequently changing affect over quick succession of time (i.e., within the same day or consecutive days, as associated with each document written by an author); as opposed to capturing over a period of time. \hartconcat performs the best across document-, wave-, and user- levels of predictions.

Empathy is much better predicted as a frequently changing human state (i.e, at the document-level) versus as an average trait for the author. However, \hartfrz is the best in predicting it as an average trait for the author (i.e., user-level).

Distress prediction showed mixed results with \hartconcat doing better for both document- and user- level predictions, while GPT-2 show same performance as \hartconcat for user-level but lower at the document-level. BERT-like models were sub-par at user-level.

\paragraph{Later Layers of Transformer-based Architectures.}
We performed initial evaluation by fetching hidden states from last layer and second-to-last layer, and comparing the different types of embeddings over the document-, wave-, and user-level tasks of predicting the human attributes (valence, arousal, empathy, and distress) (Appendix Table \ref{tab:appdx_table}). We find a dominating trend
with last layer hidden states for special tokens---LT and CLS tokens, whereas averaged token embeddings show similar results using both last and second-to-last layers with the latter having better results for auto-encoders at the document-level.  We report the results from the last layer for representations using special tokens and second-to-last layer for AT representations to use a stronger baseline auto-encoder based results in Table \ref{tab:main_results}.

\section{Conclusion}

We provided the first systematic evaluation of different types of word-, document-, and user-representations across three levels of analysis for tasks predicting both dynamically changing states and averaged trait-like user-level attributes (valence, arousal, empathy, and distress). We used traditional LMs (BERT, RoBERTA, and GPT-2) as well as human language models (versions of HaRT) and represented documents using the CLS token, the last token, the last token as \textit{insep}, U (user state from a \textit{HuLM}), and the averaged tokens hidden states for respective models. 
We hierarchically average the corresponding document representations written by a particular user to derive their user representation. Generally, document-, wave- (user over a specific time period), and user- representations based on averaged token hidden states outperform CLS, LT, \ltinsep, and U for respective models. Although the user states themselves do not prove to be a powerful representation for predicting user-level outcomes, they improve the averaged user-level performance when using user representations derived from user-state-informed tokens hidden states making HuLMs (HaRT variants) as the best performing model. 
Overall, our study and findings provide a standardized comparison and identify the best representations to select when applying LMs for psychological human-centered NLP tasks. 

\section*{Ethical Considerations}
\label{sec:ethics}
Models that incorporate the author's context and/or use data involving private and sensitive information must adopt a responsible use and release strategy. All models carry some human-level attribute information, and it is important for ethical applications (both for prosocial mental health use cases and for privacy preserving use cases) that we understand where these attributes are best represented. However, user states, user representations, or document/word representations are not mapped to any ``one'' identifiable user. These models are not trained to identify ``individual users'' or are not fed particular user-attributes, but instead use anonymized user identifiers while our evaluations compare how well they pick up on user attributes. While such models are essential for advancing research in human-centered NLP and understanding language in the context of the author, they may also pose unintended nefarious risks. For these reasons, the DS4UD data used in this study, which involves private user data, cannot be made publicly available due to ethical and privacy concerns. The affective grid circumplex tasks presented here were reviewed and approved or exempted by an academic institutional review board (IRB) as per the original dataset owners. 

\section*{Limitations}
\label{sec:limits}
The purpose of our study is to evaluate and standardize some of the commonly used Transformer-based approaches for representing users, their documents, and tokens in human-centered NLP tasks. We use some of the relevant and widely used models, exploring a few different types of representations. While our study focuses on specific models and methods, we encourage the NLP community to investigate lesser-explored approaches for modeling human context and deriving representations. We also acknowledge limitations inherent to these models and representations, such as context size and the extent of a user's historical language used as human context, which are beyond the scope for this study.

Traditional LMs and HuLMs have many different ways of deriving these representations but no study suggests the best route. And, we select representative models for these different types of LMs currently available: autoencoders (BERT-like), autoregressive (GPT2-like), and HuLMs (HaRT-like). At the same time, larger LLMs are mainly used for prompting, and not by deriving embedding-based representations to use for human-centered tasks. We use GPT-2 and BERT-like models because these are comparable in the number of parameters with HaRT-like HuLMs. Currently, larger HuLMs are not available and we encourage future studies to explore this area of research and compare the performances. 

Regarding datasets, some users and essays in the WASSA dataset were duplicates or skewed. We removed the duplicates and retained the remaining data for evaluation.
Finally, models and datasets involving sensitive user information, such as the DS4UD dataset, require careful and responsible usage. Due to privacy concerns, the authors of these datasets cannot make them publicly available. The datasets for human-centered NLP tasks are often small-scale due to a variety of reasons including: (i) availability of user information, (ii) privacy issues, (iii) difficulty in collecting labeled datasets at the user level for a continued period of time. DS4UD dataset provides a rich historical language of authors (essays) collected for a continued period of time (e.g., DS4UD was collected over 3 years, from early 2021 until 2024), with associated multiple human-level attributes. Thus, this provides us with an opportunity to analyze the stable traits and changing human states of people. To the best of our knowledge, no studies provide rich longitudinal data at such scale. WASSA datasets are publicly available and provide rich textual data with associated labels for multiple documents written by a particular author, thus enabling the prediction of different empathetic (or distress) states of a person. To sum up, these datasets provide complex scenarios encapsulating changing human states and averaged traits of a person. More importantly, DS4UD also provides the temporally changing human-level attributes with self-reported labels instead of inferred attributes or annotated labels. Furthermore, these datasets are essay-like language which reduces the limitations of social-media datasets that are short and not necessarily continuous over time.

\section*{Acknowledgments}
This research is supported in part by the Office of the Director of National Intelligence (ODNI), Intelligence Advanced Research Projects Activity (IARPA), via the HIATUS Program contract \#2022-22072200005, and a grant from the NIH-NIAAA (R01 AA028032). The views and conclusions contained herein are those of the authors and should not be interpreted as necessarily representing the official policies, either expressed or implied, of ODNI, IARPA, any other government organization, or the U.S. Government. The U.S. Government is authorized to reproduce and distribute reprints for governmental purposes notwithstanding any copyright annotation therein.


\bibliography{anthology, custom}

\appendix

\section{Supplementary Results}

\begin{table*}[]
    \centering
    \begin{small}
    \begin{tabular}{l|cccccccc|cccc}
        
        \hline
        
        \tbf{} & \multicolumn{12}{c}{\tbf{Dynamic States}} \\
        
        \tbf{Method} & \multicolumn{8}{c|}{\tbf{Document-Level}} & \multicolumn{4}{c}{\tbf{Wave-Level}} \\
        
         \tbf{} & \multicolumn{2}{c|}{\tbf{Val}} & \multicolumn{2}{c|}{\tbf{Aro}} & \multicolumn{2}{c|}{\tbf{Emp}} & \multicolumn{2}{c}{\tbf{Dis}} & \multicolumn{2}{c|}{\tbf{Val}} & \multicolumn{2}{c}{\tbf{Aro}} \\

         \tbf{} & \tbf{L} & \tbf{SL} & \tbf{L} & \tbf{SL} & \tbf{L} & \tbf{SL} & \tbf{L} & \tbf{SL} & \tbf{L} & \tbf{SL} & \tbf{L} & \tbf{SL} \\
         
         \hline

         GPT-2 &&&&&&&&&&&&\\
         
         \hspace{15pt} LT &\textbf{0.53}&0.48&\textbf{0.25}&0.18&0.71&0.71&0.58&0.58&\textbf{0.64}&0.61&\textbf{0.28}&0.15 \\
         
         \hspace{15pt} AT &0.60&0.60&0.34&0.34&\textbf{0.70}&0.69&\textbf{0.58}&0.54&0.76&\textbf{0.77}&0.29&\textbf{0.33} \\

         BERT &&&&&&&&&&&&\\
         
         \hspace{15pt} CLS & 0.58&\textbf{0.59}&\textbf{0.30}&0.29&0.70&\textbf{0.71}&\textbf{0.66}&0.60&\textbf{0.74}&0.70&0.28&\textbf{0.29} \\
         
         \hspace{15pt} AT &0.61&0.61&0.35&0.35&0.71&\textbf{0.72}&0.59&0.59&\textbf{0.77}&0.75&\textbf{0.29}&0.26 \\

         RoBERTa &&&&&&&&&&&&\\
         
         \hspace{15pt} CLS &\textbf{0.62}&0.59&\textbf{0.35}&0.30&\textbf{0.77}&0.69&\textbf{0.63}&0.42&\textbf{0.75}&0.71&0.29&\textbf{0.30} \\
         
         \hspace{15pt} AT &0.63&0.63&0.36&0.36&0.71&\textbf{0.72}&0.60&\textbf{0.63}&\textbf{0.79}&0.75&0.33&0.33 \\

         \hartfrz &&&&&&&&&&&& \\
         
         \hspace{15pt} LT &\textbf{0.52}&0.51&\textbf{0.22}&0.20&0.70&0.70&\textbf{0.57}&0.56&\textbf{0.69}&0.68&\textbf{0.21}&0.16\\
         
         \hspace{15pt} \ltinsep &0.55&\textbf{0.56}&\textbf{0.23}&0.21&\textbf{0.76}&0.74&\textbf{0.67}&0.52&0.69&0.69&\textbf{0.22}&0.20 \\

         \hspace{15pt} AT &0.63&\textbf{0.64}&0.33&0.33&\textbf{0.77}&0.75&\textbf{0.70}&0.68&0.79&0.79&0.27&0.27 \\

         \hspace{15pt} U &\multicolumn{2}{c}{-}&\multicolumn{2}{c}{-}&\multicolumn{2}{c}{-}&\multicolumn{2}{c|}{-}&\multicolumn{2}{c}{-}&\multicolumn{2}{c}{-} \\


         \hartconcat &&&&&&&&&&&& \\         
         
         \hspace{15pt} LT &\textbf{0.53}&0.52&\textbf{0.22}&0.18&\textbf{0.72}&0.70&\textbf{0.62}&0.60&0.70&0.70&\textbf{0.16}&0.13 \\
         
         \hspace{15pt} \ltinsep &0.57&\textbf{0.59}&0.26&0.26&\textbf{0.76}&0.46&\textbf{0.56}&0.48&\textbf{0.75}&0.74&\textbf{0.30}&0.28 \\

         \hspace{15pt} AT &0.64&\textbf{0.65}&0.34&0.34&\textbf{0.77}&0.75&\textbf{0.72}&0.70&0.80&0.80&\textbf{0.30}&0.27 \\

         \hspace{15pt} U &\multicolumn{2}{c}{-}&\multicolumn{2}{c}{-}&\multicolumn{2}{c}{-}&\multicolumn{2}{c|}{-}&\multicolumn{2}{c}{-}&\multicolumn{2}{c}{-} \\

         \hartonedoc &&&&&&&&&&&& \\
         
         \hspace{15pt} LT &\textbf{0.57}&0.53&\textbf{0.27}&0.21&\textbf{0.68}&0.63&\textbf{0.56}&0.52&\textbf{0.72}&0.69&\textbf{0.17}&0.13 \\
         
         \hspace{15pt} \ltinsep &0.64&\textbf{0.65}&\textbf{0.34}&0.33&\textbf{0.74}&0.73&\textbf{0.66}&0.65&0.79&\textbf{0.80}&\textbf{0.27}&0.25 \\

         \hspace{15pt} AT &0.63&0.63&0.36&0.36&\textbf{0.74}&0.73&\textbf{0.64}&0.63&0.80&0.80&\textbf{0.29}&0.28 \\

         \hspace{15pt} U &\multicolumn{2}{c}{0.52}&\multicolumn{2}{c}{0.15}&\multicolumn{2}{c}{0.46}&\multicolumn{2}{c|}{0.40}&\multicolumn{2}{c}{0.65}&\multicolumn{2}{c}{0.01} \\
         
         \hline
         
    \end{tabular}
    \end{small}
    \caption{We evaluate the performance differences in hidden states from the last layer (``L'') and second-to-last layer (``SL'') for all models. We find second to last layer to perform better for AT hidden states in case of auto-encoders for document-level tasks, and last layer to work generally well for the other types of representations across models. AT embeddings from L and SL show similar results for most other outcomes. Bold indicates better performance between L and SL for each model/representation/outcome.}
    \label{tab:appdx_table}
\end{table*}

\begin{table*}[]
    \centering
    \begin{small}
    \begin{tabular}{l|cccccccc}
        
        \hline
        
        \tbf{} & \multicolumn{8}{c}{\tbf{Trait-like}} \\
        
        \tbf{Method} & \multicolumn{8}{c}{\tbf{User-Level}} \\
        
         \tbf{} & \multicolumn{2}{c|}{\tbf{Val}} & \multicolumn{2}{c|}{\tbf{Aro}} & \multicolumn{2}{c|}{\tbf{Emp}} & \multicolumn{2}{c}{\tbf{Dis}} \\

         \tbf{} & \tbf{L} & \tbf{SL} & \tbf{L} & \tbf{SL} & \tbf{L} & \tbf{SL} & \tbf{L} & \tbf{SL} \\
         
         \hline

         GPT-2 &&&&&&&&\\
         
         \hspace{15pt} LT &\textbf{0.63}&0.52&-0.14&\textbf{-0.01}&\textbf{0.37}&0.36&0.42&\textbf{0.47} \\
         
         \hspace{15pt} AT &0.75&0.75&\textbf{0.19}&0.18&0.52&0.52&0.64&\textbf{0.68} \\

         BERT &&&&&&&&\\
         
         \hspace{15pt} CLS & \textbf{0.71}&0.68&\textbf{0.09}&0.05&0.45&\textbf{0.48}&0.42&0.42 \\
         
         \hspace{15pt} AT &\textbf{0.75}&0.73&\textbf{0.17}&0.14&0.44&\textbf{0.45}&0.37&\textbf{0.39} \\

         RoBERTa &&&&&&&&\\
         
         \hspace{15pt} CLS &\textbf{0.72}&0.61&0.10&\textbf{0.13}&\textbf{0.51}&0.48&0.59&\textbf{0.66} \\
         
         \hspace{15pt} AT &0.75&0.75&\textbf{0.20}&0.16&\textbf{0.62}&0.59&\textbf{0.58}&0.53 \\

         \hartfrz &&&&&&&& \\
         
         \hspace{15pt} LT &\textbf{0.73}&0.72&\textbf{0.18}&0.13&0.42&\textbf{0.43}&\textbf{0.60}&0.59\\
         
         \hspace{15pt} \ltinsep &\textbf{0.74}&0.73&\textbf{0.23}&0.20&\textbf{0.45}&0.44&\textbf{0.62}&0.47 \\

         \hspace{15pt} AT &0.72&\textbf{0.73}&0.35&0.35&\textbf{0.68}&0.67&0.64&\textbf{0.66} \\

         \hspace{15pt} U &\multicolumn{2}{c}{0.59}&\multicolumn{2}{c}{-0.01}&\multicolumn{2}{c}{0.46}&\multicolumn{2}{c}{0.49} \\


         \hartconcat &&&&&&&& \\         
         
         \hspace{15pt} LT &\textbf{0.74}&0.71&\textbf{0.18}&0.12&0.42&\textbf{0.43}&\textbf{0.68}&0.63 \\
         
         \hspace{15pt} \ltinsep &\textbf{0.77}&0.75&\textbf{0.15}&0.13&0.61&\textbf{0.65}&0.55&\textbf{0.59} \\

     \hspace{15pt} AT &0.76&\textbf{0.77}&\textbf{0.35}&0.34&\textbf{0.64}&0.62&0.63&\textbf{0.65} \\

         \hspace{15pt} U &\multicolumn{2}{c}{0.65}&\multicolumn{2}{c}{0.06}&\multicolumn{2}{c}{0.59}&\multicolumn{2}{c}{0.49} \\

         \hartonedoc &&&&&&&& \\
         
         \hspace{15pt} LT &\textbf{0.72}&0.68&\textbf{0.24}&0.23&0.41&\textbf{0.42}&0.55&\textbf{0.56} \\
         
         \hspace{15pt} \ltinsep &\textbf{0.72}&0.71&0.09&0.09&\textbf{0.40}&0.39&\textbf{0.50}&0.47 \\

         \hspace{15pt} AT &0.75&0.75&\textbf{0.18}&0.12&0.57&\textbf{0.58}&\textbf{0.52}&0.51 \\

         \hspace{15pt} U &\multicolumn{2}{c}{0.66}&\multicolumn{2}{c}{0.04}&\multicolumn{2}{c}{0.50}&\multicolumn{2}{c}{0.47} \\
         
         \hline
         
    \end{tabular}
    \end{small}
    \caption{We evaluate the performance differences in hidden states from the last layer (``L'') and second-to-last layer (``SL'') for all models. AT embeddings from L and SL show mixed and similar results for most outcomes at user-level. Last layer works generally well for the other types of representations across models. Bold indicates better performance between L and SL for each model/representation/outcome.}
    \label{tab:appdx_table2}
\end{table*}

\end{document}